\documentclass[letterpaper, 10 pt, conference]{ieeeconf}  

\IEEEoverridecommandlockouts                              

\usepackage{graphics} 
\usepackage{epsfig} 
\usepackage{amsmath} 
\usepackage{amssymb}  
\usepackage{mathtools}  
\usepackage{float}
\usepackage{verbatim}
\usepackage{textcomp}
\usepackage{siunitx}
\usepackage{subcaption}
\usepackage{cite}
\usepackage{hyperref}
\usepackage{siunitx}
\usepackage{color}
\hypersetup{
    colorlinks=true,
    linkcolor=blue,
    filecolor=magenta,      
    urlcolor=blue,
    pdftitle={Overleaf Example},
    pdfpagemode=FullScreen,
    }

\usepackage{geometry}
\DeclareMathOperator*{\argmin}{arg\,min}  

\definecolor{kkBlue}{rgb}{0,0.0,0.0}
\definecolor{kkBlack}{rgb}{0,0.0,0.0}
\newcommand{\kk}[1]{\textcolor{kkBlack}{#1}}
\newcommand{\ky}[1]{\textcolor{kkBlue}{#1}}
\newcommand{\mc}[1]{\textcolor{black}{#1}}

\definecolor{mmRed}{rgb}{0.0,0.0,0.0}
\newcommand{\mm}[1]{\textcolor{mmRed}{#1}}

\definecolor{mmPurple}{rgb}{0.0,0.0,0.0}

\definecolor{mmRed2}{rgb}{0.0,0.0,0.0}
\newcommand{\mmedit}[1]{\textcolor{mmRed2}{#1}}

\definecolor{mmRed3}{rgb}{0.0,0.0,0.0}
\newcommand{\mmledit}[1]{\textcolor{mmRed3}{#1}}

\title{\LARGE \bf
\kk{Robust Visual Teach and Repeat for UGVs Using 3D Semantic Maps}}

\author{Mohammad Mahdavian, KangKang Yin and Mo Chen
\thanks{The authors are with School of Computing Science, Simon Fraser University, Burnaby, Canada
        {\tt\small \{mmahdavi, kkyin, mochen\}@sfu.ca}}%
}

\newcommand\norm[1]{\left\lVert#1\right\rVert}

\renewcommand{\baselinestretch}{0.92}

\begin{document}

\maketitle
\thispagestyle{empty}
\pagestyle{empty}

\begin{abstract}

\kk{We propose a Visual Teach and Repeat (VTR) algorithm using semantic landmarks extracted from environmental objects for ground robots with fixed mount monocular cameras. 
The proposed algorithm is robust to changes in the starting pose of the camera/robot, where a pose is defined as the planar position plus the orientation around the vertical axis. 
VTR consists of a teach phase in which a robot moves in a prescribed path, and a repeat phase in which the robot tries to repeat the same path starting from the same or a different pose. 
Most available VTR algorithms are pose dependent and cannot perform well in the repeat phase when starting from an initial pose far from that of the teach phase. 
\ky{To achieve more robust pose independency, the key is to generate a 3D semantic map of the environment containing the camera trajectory and the positions of surrounding objects during the teach phase. For specific implementation, we use ORB-SLAM to collect the camera poses and the 3D point clouds of the environment, and YOLOv3 to detect objects in the environment. We then combine the two outputs to build the semantic map.} In the repeat phase, we relocalize the robot based on the detected objects and the stored semantic map. The robot is then able to move toward the teach path, and repeat it in both forward and backward directions. 
\ky{We have tested the proposed algorithm in different scenarios and compared it with two most relevant recent studies. \mmedit{Also, we compared our algorithm with two image-based relocalization methods. One is purely based on ORB-SLAM and the other combines Superglue and RANSAC}}. 
The results show that our algorithm is much more robust with respect to pose variations as well as environmental alterations. Our code and data are available at the following Github page: \url{https://github.com/mmahdavian/semantic_visual_teach_repeat}.}

\end{abstract}


\section{Introduction}

\kk{Visual Teach and Repeat (VTR) is an important task in robotic navigation. It has practical applications for repetitive tasks such as surveillance and transportation, especially for robots that are not equipped with GPS sensors, or in indoor areas where GPS sensors have low accuracy. VTR is thus an alternative for navigating an Unmanned Ground or Aerial Vehicle (UGV or UAV) at a relatively low cost, as it only requires a normal monocular camera. 
}

\begin{figure}[t]
      \centering
      \includegraphics[scale=0.04]{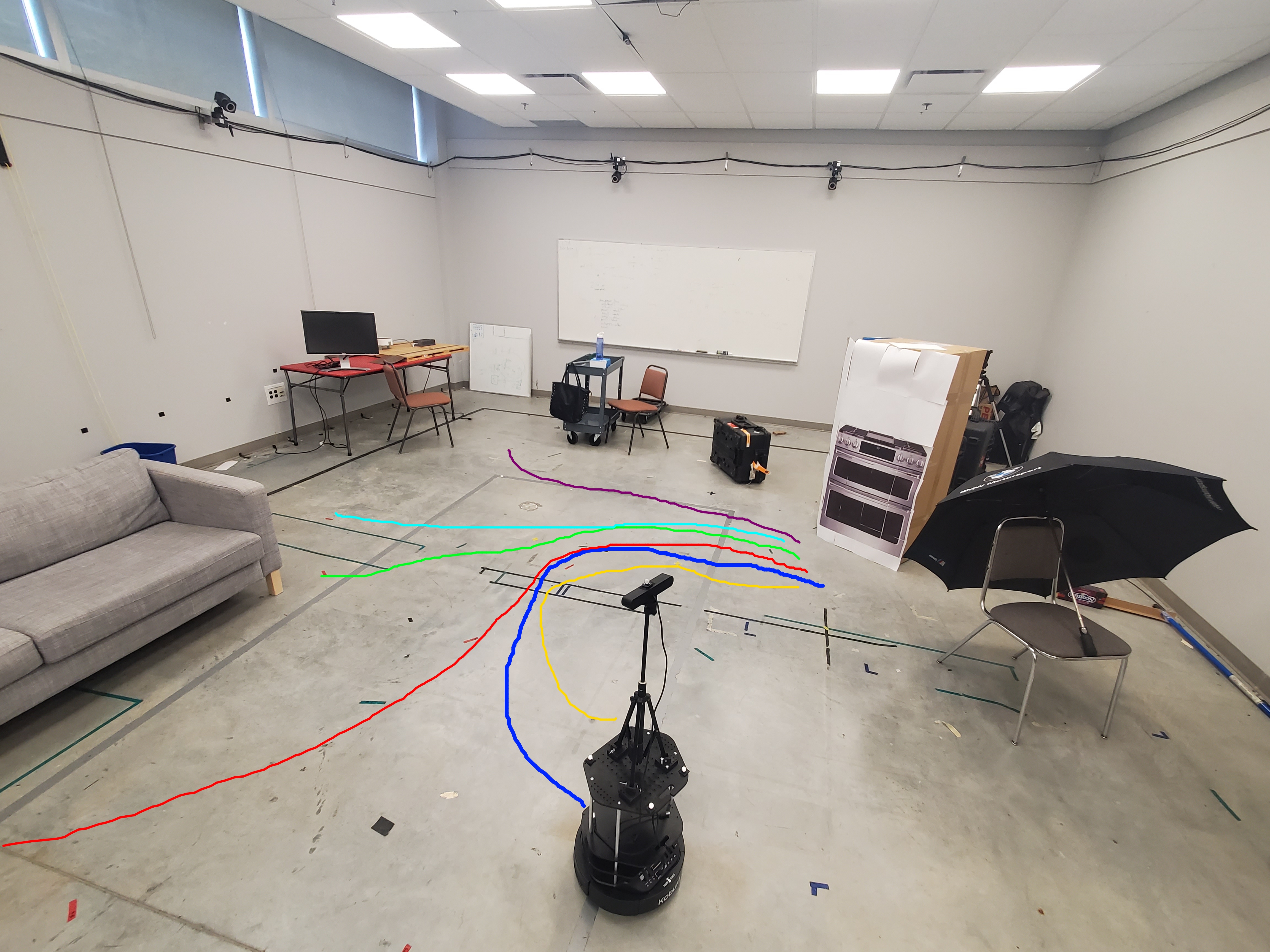}  
      \caption{An example of VTR paths starting from different locations and orientations.}
      \label{teach_repeat}
   \end{figure}

\kk{VTR consists of a teach phase and a repeat phase. In the teach phase, the robot is driven by a user, or a path planner, and captures images along and around the path. During the repeat phase, the robot tries to repeat the same path starting from arbitrary locations and orientations, using only images captured by the camera. \ky{Fig.~\ref{teach_repeat} demonstrates a teach path and multiple repeat paths using our robot in our testing area.}}

\kk{There are two key challenges for VTR algorithms. First, during the repeat phase depending on the starting pose, the onboard camera may not be looking from the same location and angle as during the teach phase. In extreme cases such as transportation robots working in mines, the robot may need to move back to the starting point, so the camera position and orientation both change dramatically. Second, given that onboard memory is usually limited and robots cannot store all the raw images of the teach phase, VTR algorithms are desired to abstract the raw data to reduce storage as well as to relocalize in an online fashion.}

\kk{The two common ways to process and abstract raw images for VTR are 1) creating 2D or 3D SLAM maps and 2) extracting \ky{and memorizing} local features such as SIFT\cite{c1}, SURF\cite{c2}, BRIEF\cite{c3} or ORB\cite{c4} for each frame. The first way, creating accurate SLAM maps, requires advanced sensors such as Lidar, RGBD, or stereo cameras\mmedit{~\cite{c10}}, which limits the application of the VTR algorithms. \mm{As an extreme version of VTR, we target UGVs that are only equipped with monocular cameras as a low-cost, light-weight, and ubiquitous sensor which can be mass-produced and quickly deployed. The SLAM maps generated using the monocular cameras are not highly accurate and in some cases scale ambiguous\cite{c6}. Therefore, we cannot solely rely on them to relocalize a camera in between any two runs.}}
\kk{The second way, using local image features, only requires monocular cameras, but is problematic when the camera pose changes dramatically between the teach and repeat phases. Previous work using local features alone reported failures when viewing angles differ more than approximately 30 degrees \cite{c5,c6,c7}.}

\kk{\textbf{Contributions}: To the best of our knowledge, \mm{in the domain of VTR algorithms,} we are the first to successfully use monocular cameras to obtain 3D semantic maps of the environmental objects to achieve more robust VTR. During the teach phase, we combine both SLAM and local image features to create 3D semantic maps of the environment. The maps contain both object locations and semantic labels, together with the 3D camera poses on the teach path. In the repeat phase, our relocalization algorithm uses recognized objects as reference landmarks to align the camera poses to those in the teach map. Therefore the relocalization algorithm is robust with respect to starting camera pose changes, including planar position differences and the viewing angle variations, \mm{without requiring an accurate SLAM map}. In addition, object movements between teach and repeat phases can be tolerated to a certain extent. Our results show that the repeat phase can start from a large range of initial poses in the environment, far away from the teach phase initial position with large viewing angle differences. \mmedit{This was not achievable by previous methods including using visual SLAM and image-based relocalization methods such as the combination of Superglue \cite{c26} and RANSAC \cite{c27}}. We also demonstrate the ability to repeat the teach path in backward directions, which is not possible by any previous VTR work due to significant pose changes.}

\section{Related Works}

\kk{VTR has been investigated in a few studies for UAVs\cite{c8,c9} and UGVs\cite{c10}, most of which are based on local image features. As one of the earliest studies, Furgale and Barfoot\cite{c11} built a manifold map of overlapping submaps as the robot was piloted along a route. The map was then used for localization as the robot repeats the route autonomously. Barfoot later developed a method called Multi-Experience Localization (MEL) by using local image features\cite{c12,c13,c14,c15,c16,c17}. Such methods boost the performance by bridging between local image features found in multiple repeats.  All these methods are sensitive to the initial viewing angle. The robot is possible to lose track of the path as the locations of local image features change drastically according to the viewpoint.}

\kk{\ky{More recently}, Camara et al. trained deep learning models to solve VTR with high end-point accuracy~\cite{c18}. They used a CNN (Convolutional Neural Network) model as a feature extractor to build a feature database during the teach phase. At the repeat phase, the extracted features by the CNN model were compared with the feature database for visual place recognition. A horizontal offset estimator was also used to find the direction that the robot needs to move toward. However, the algorithm is still not robust to viewpoint changes, and may lose track of the path when starting from a location far away from the actual teach path.}

\ky{Most recently, Dall'Osto et al. developed a VTR algorithm that utilized odometry information,  paired with a correction signal driven by a computationally lightweight visual processing module~\cite{c25}. Their algorithm cannot handle extreme initial pose changes, as will be shown in our comparative study.}

\kk{In order to alleviate the pose-dependence problem, Ghasemi Toudeshki et al. proposed to utilize environmental objects~\cite{c19}, which can be robust semantic features of the environment that are viewpoint independent. Their algorithm memorized the semantic objects found during the teach phase for a UAV. Later a Seq-SLAM-like relocalization module used these objects to find the correct path toward the end-point. The work used objects as 2D \ky{image features without creating a 3D map}, which resulted in inferior repeat accuracy compared to ours that uses environmental objects as \ky{3D features. Again, we will show comparisons with this algorithm in our results section.}}

In order to solve the relocalization part of the VTR, there are other image-based algorithms available. For example, the combination of Superglue~\cite{c26} and RANSAC~\cite{c27} can be used for this purpose. Superglue is an image feature matching method based on a neural network that matches two sets of local features by jointly finding correspondences and rejecting non-matchable points. We combine it with RANSAC to find a transformation between teach and repeat phases initial poses. Also, ORB-SLAM\cite{c6} has relocalization capability. We will compare our algorithm with these two methods as well.

\kk{Our method shares the same motivation as~\cite{c19} to use environmental objects as semantic features, but for the VTR problem of UGVs. Moreover, the 3D object locations are used to increase the accuracy of the algorithm, in addition to decreasing the sensitivity to initial robot poses. We were \ky{partly inspired by camera relocalization methods described in \cite{c5} and \cite{c24}, which demonstrated that semantic objects can make camera location estimation more accurate and robust to significant viewing angle changes. As these methods were only applied to relocalization and not VTR, we did not perform a direct comparison with them. We note that our relocalization algorithm simplifies the one in \cite{c5} by using only 3D positions of the semantic objects. It is faster and more practical for VTR applications, and enables robust relocalization for even backward repeats not possible before. \mmedit{Although there are other possible methods for object matching like Hungarian algorithm~\cite{c5} or RANSAC~\cite{c27}, we took a simple approach as it already works well for our case. }}}

\begin{figure}[t]
\centering
\includegraphics[scale=0.2]{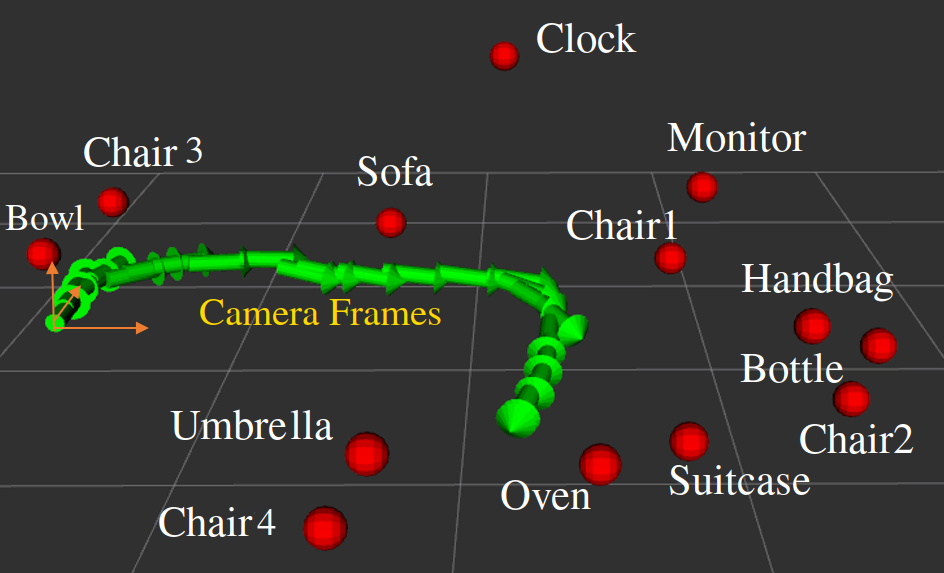} 
\caption{\kk{An example of semantic map created in teach phase containing semantic object labels and positions as well as camera keyframes. Red dots represent the upper middle point of the corresponding objects.}}
\label{semantic_map}
\end{figure}

\section{Preliminaries}

\subsection{Overview of Proposed Method}

\kk{We propose a VTR algorithm that is robust to large viewpoint changes, e.g., when the robot starts from a position far away from the teach path. \ky{The key of our method is to generate a semantic 3D map\mm{, rather than a complete SLAM occupancy grid map,} from the environment containing observed objects as well as the camera trajectory. We use a SLAM module to track the camera path and a recognition module to detect objects from image frames. As an example implementation, we choose the well-known ORB-SLAM~\cite{c6} to obtain camera poses from local image features, and a CNN-based model You Only Look Once (YOLOv3)~\cite{c20} to recognize objects in the environment. Note that other choices may work within our framework as well. We combine the outputs of ORB-SLAM and YOLOv3 to build a 3D semantic map}, as shown in Fig.~\ref{semantic_map}, that contains semantic object labels, 3D object positions, along with the robot trajectory.}

\subsection{ORB-SLAM}
\label{sec:ORB-SLAM}

\kk{ORB-SLAM~\cite{c6} has been widely adopted to reconstruct the camera trajectory and generate a sparse 3D reconstruction of the environment for monocular, stereo, and RGB-D cameras in real-time. It is also able to perform map reuse, loop closure, and relocalization with small viewing angle changes. In this work, we use ORB-SLAM for a monocular camera. The SLAM module gets initialized after receiving multiple image frames when the camera starts moving. The bundle adjustment algorithm takes in the motion of extracted 2D local ORB features and produces a map containing camera poses and 3D positions of these ORB features. Fig.~\ref{yolo_orb} shows an example of the 2D local ORB features detected during a robot motion. The origin of the generated map is the starting point where the ORB-SLAM module was initialized. However, the generated map is ambiguous in its scale. Even ORB-SLAM maps for the same area may have different scales in different runs. We will discuss in Section~\ref{sec:relocalization} how to handle this scale ambiguity in our relocalization algorithm.}

\mmedit{We cannot solely rely on ORB-SLAM or any other similar visual SLAM algorithms to relocalize the camera in between any two different poses due to deficiency of similar local image features after dramatic view angle changes, \mmedit{as shown in Section \ref{sec:comparison}}. Therefore, we combine local features obtained from visual SLAM with an object detector to add the necessary semantic information for robust VTR.}

\subsection{YOLOv3}
\label{sec:YOLOv3}

\begin{figure}[t]
  \centering
  \includegraphics[scale=0.33]{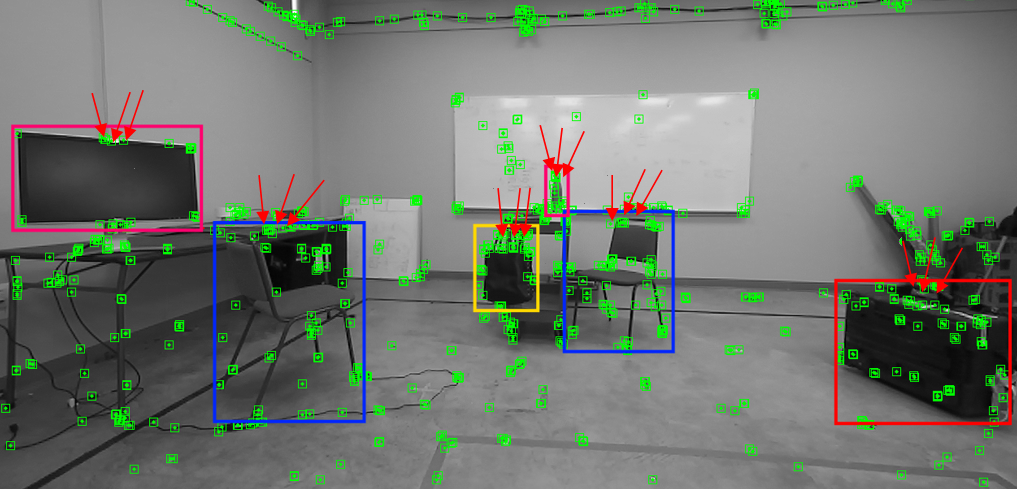}  
  \caption{\kk{Local features detected by ORB-SLAM and bounding boxes detected by YOLOv3. \mm{The closest image features within a threshold} to the upper middle point of the bounding boxes, pointed by the red arrows, are used for calculating the 3D positions of the corresponding objects.}}
  \label{yolo_orb}
\end{figure}

\kk{To detect objects in each image frame, we employ a CNN-based model, YOLOv3~\cite{c20}. The YOLOv3 model applies a single neural network to the full image to extract image features first. Then it divides the image into several regions and predicts the location of the bounding box, the semantic label, and their confidence scores for each object. To achieve high accuracy, the network was trained on 24 most common objects in indoor areas, such as TV-Monitor, Sofa, Chair, Umbrella, Clock, Bottle, etc., from the COCO 2017 dataset. Also, the Darknet-ROS~\cite{c21} package was utilized to publish objects detected by YOLOv3 to ROS. Fig.~\ref{yolo_orb} shows the detected objects by YOLOv3 during a robot motion.}

\mc{YOLOv3 was chosen for its high accuracy as well as its fast inference speed, which is necessary since several modules need to run in parallel.
In general, any other highly accurate and fast object detector could be used for our VTR algorithm.
}

\section{Teach Phase}

\kk{In the teach phase of VTR algorithms, the robot moves either with manual control from a user, or by a path planning algorithm. We call the trajectory of this motion the {\it teach path}, which the robot needs to memorize in order to repeat it later. To this end, we wish to create a 3D semantic map, referred to as the {\it teach map} hereafter, that contains camera keyframes, semantic object labels, and object positions in 3D. Camera keyframes can be obtained from ORB-SLAM directly, as described in Section~\ref{sec:ORB-SLAM}. Simultaneously, a YOLOv3 model is utilized to detect and classify objects in the streaming images to provide us the semantic object labels.}

\kk{In order to obtain 3D object locations, we combine the 3D positions of ORB features from ORB-SLAM and the estimated object bounding boxes from YOLOv3. More specifically, we estimate the 3D object position from the 3D positions of the ORB features inside the corresponding object's bounding box. Empirically, the upper middle area of each object is the least occluded and most visible region of the object in image frames. Therefore, we first find the closest ORB features to the upper-middle point of the object bounding box within a threshold, and then use their average 3D positions as the object location in the semantic map. \mm{A similar heuristics was implemented in \cite{c5} to create a 3D semantic map of the environment to relocalize the camera poses.}
In Fig.~\ref{yolo_orb}, we use red arrows to point out the chosen ORB feature points for estimating 3D object locations.} 

\kk{Object locations estimated from different image frames are usually not identical numerically, even for the same object, due to noise and errors in camera movement and image feature detection. We therefore only add a semantic landmark and its object location into the semantic map when its estimated 3D position is beyond a reasonable threshold away from any previously added objects with a similar label. We also ignore bounding boxes that are only partially observable inside the image frames to avoid large estimation errors.} 

\mc{For simplicity, we ignore any duplicate objects during the teach phase and only keep track of unique objects in the semantic map. In principle, duplicate object classes can be considered, but could introduce unnecessary ambiguities in the semantic map. In practice, many unique objects exist in most indoor environments. In addition, as we demonstrate in Section \ref{sec:expt}, a very sparse semantic map containing 3D positions of just a few objects is sufficient for robust VTR.}

\kk{By the end of the teach phase, we have obtained a semantic teach map containing 3D positions of environmental objects as well as their semantic labels, and a camera pose trajectory $P^k$ where $k \in \{1,\cdots,K\}$ is the keyframe index. Fig.~\ref{semantic_map} shows a sample semantic map together with the origin of the map for our testing scene used in the experiments.}

\section{Repeat Phase}

\kk{We aim to develop a repeat algorithm that is highly robust to large variations of the robot starting poses, such as faraway starting locations or viewing angles opposite to the ones of the teach phase. Tolerance to reasonable environment changes, such as relocation of a subset of objects, is desirable as well. The algorithm should also be able to repeat the teach path with reasonable accuracy, relative to the size of the robot.} 

\kk{The key to such a robust VTR algorithm is to accurately relocalize the robot pose with respect to the teach map coordinate frame. For this purpose, we developed an optimization-based relocalization algorithm that finds the \textbf{best matching pair of objects} between the teach and repeat maps to relocalize the robot robustly and accurately. After relocalization, the robot can simply move toward the closest point on the teach path and repeat it to the end or move back to the starting point.}

\subsection{Relocalization}
\label{sec:relocalization}

\kk{To relocalize the robot in the repeat phase, 3D positions of objects in the environment estimated during the teach phase are used as reference landmarks to transform the robot pose in repeat map $P_{r}$ to its counterpart in the teach map $P_{t}$. We note that the estimated 3D positions of the objects contain noise and errors from various sources, due to estimation errors from ORB-SLAM, occasional object relocation in the environment, and noise caused by the movement of the robot. Therefore, we robustly estimate the relative transformation from $P_{r}$ to $P_{t}$ using the best matching pair of objects in the environment. Furthermore, we assume there are multiple unique objects in the environment. We only consider unique objects to reduce ambiguity caused by repetitive objects, and leave it as future work to utilize repetitive objects for relocalization.} 

\kk{In general, there are 6 Degrees of Freedom (DoFs) for a given rigid body. However, our camera is mounted with a fixed base on a UGV that can only move in the horizontal ground plane and rotate around the vertical axis. Therefore, there are only 3 DoFs left for the robot/camera. In addition, ORB-SLAM outputs are ambiguous in scale. Thus there are in total four unknowns ($s,\delta x,\delta y,\delta\theta$) in the relative transformation between two semantic maps, where $s$ is the scalar, ($\delta x$,$\delta y$) are the planar translation, and $\delta\theta$ is the rotation around the vertical axis. We thus require the positions of two matching environmental objects to solve for the four unknowns. The scalar $s$ can be easily calculated by dividing the relative distances of the pair of objects in the two maps. We now detail how to solve for the relative transformation between $P_{r}$ and $P_{t}$ from intermediate coordinate frames estimated from the two matching objects.} 

\begin{figure}[t]
\centering
\includegraphics[scale=0.28]{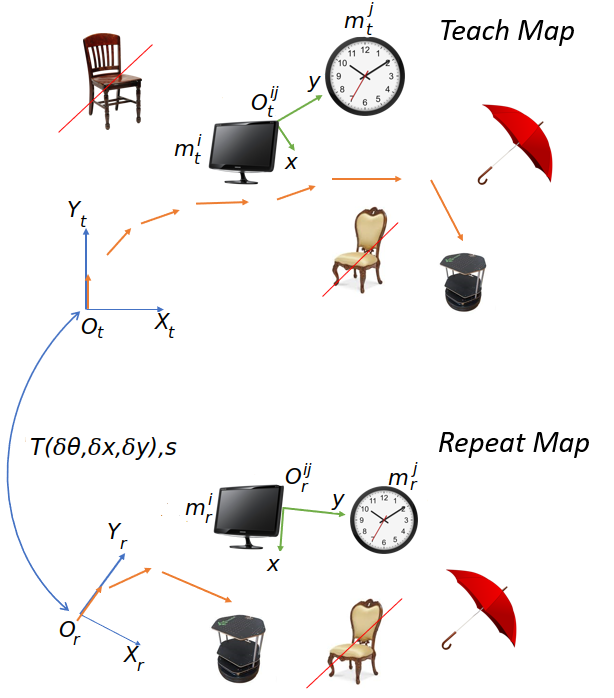} 
\caption{\kk{Relocalization between the teach and repeat phase calculates $\delta x, \delta y , \delta\theta$ and $s$ from two matching objects.}}
\label{translation}
\end{figure}

\kk{Fig.~\ref{translation} illustrates various coordinate frames involved in relocalization, together with some environmental objects. $O_t$ and $O_r$ are the two 2D coordinate frames automatically extracted by ORB-SLAM from the robot starting pose in teach and repeat phases, respectively. They define the map's origin and starting orientation. We first ignore repetitive objects in both maps, such as the chairs in Fig.~\ref{translation}. \ky{Then object pairs with the same labels in both maps are enumerated, e.g., the monitor and the clock.} Hereafter we denote individual objects as $m^{i}$ and $m^{j}$, and the corresponding object pair as $m^{ij}$. Next, a new 2D coordinate frame $O^{ij}$ is defined from the chosen objects $m^{ij}$ for both maps. The $y$-axis is the vector pointing from one object to the other in the horizontal plane, e.g., the vector from the monitor to the clock. The $x$-axis is the cross product of the $y$-axis and the up direction. We denote a 3-DoF transformation matrix $T$ with one rotational DoF $\alpha$ and two translational DoFs $dx,dy$ as in \eqref{transformation}. Then, the transformation of all objects from a map coordinate frame, either $O_{t}$ or $O_{r}$, to its corresponding object coordinate frame, either $O^{ij}_t$ or $O^{ij}_r$, is denoted by \eqref{teach_to_mi} and \eqref{repeat_to_mi}.} 
\vspace{-8pt}

\begin{equation} \label{transformation}
T(\alpha,dx,dy) =  \begin{bmatrix}
\cos \alpha & -\sin \alpha & 0 & dx\\
\sin \alpha &  \cos \alpha & 0 & dy\\
0 & 0 & 1 & 0\\
0 & 0 & 0 & 1\\
\end{bmatrix}
\end{equation}
\vspace{-12pt}

\begin{equation}\label{teach_to_mi}
T(\alpha_t,dx_t,dy_t) = \text{}^{O^{ij}_t}_{O_{t}}T
\end{equation}
\vspace{-12pt}

\vspace{-4pt}
\begin{equation}\label{repeat_to_mi}
T(\alpha_r,dx_r,dy_r) = \text{}^{O^{ij}_r}_{O_{r}}T
\end{equation}
\vspace{-6pt}

\noindent \kk{where the subscript $_t$ and $_r$ denote the various quantities in either the teach or repeat map. We denote the coordinates of all objects in a map as $M_t=\{m^k_t, k \in (1 \cdots N_t)\}$ or $M_r=\{m^k_r, k \in (1 \cdots N_r)\}$. Then, using \eqref{teach_map_to_mi} and \eqref{repeat_map_to_mi} we can transform an object $m^k_t$ in $M_t$ or $m^k_r$ in $M_r$, in their original map coordinates to their corresponding object coordinates as follows:} \begin{equation}\label{teach_map_to_mi}
{}^{O^{ij}_t}m^k_t = \text{}^{O^{ij}_t}_{O_{t}}T \times {}^{O_t}m^k_t  
\end{equation}
\vspace{-15pt}

\begin{equation}\label{repeat_map_to_mi}
{}^{O^{ij}_r}m^k_r = \text{}^{O^{ij}_r}_{O_{r}}T \times {}^{O_r}m^k_r
\end{equation}

\vspace{-3pt}
\kk{Then the aggregated error $\gamma$ between positions of all $N$ unique objects that appear in both maps can be calculated:}

\vspace{-8pt}
\begin{equation}
\label{eq-gamma}
\gamma(O^{ij}_t,M_t,O^{ij}_r,M_r,s^{ij}) = 
\displaystyle\sum\limits_{k=1}^{N}\left( {}^{O^{ij}_t}m^k_t-s^{ij}\times {}^{O^{ij}_r}m^k_r \right)^2
\end{equation}

\kk{$s^{ij}$ is a scale factor introduced by ORB-SLAM, which can be estimated by:}
\vspace{-10pt}

\begin{equation} 
\label{eq_scale}
s^{ij} = \norm{\overrightarrow{m^{ij}_t}} \Big / \norm{\overrightarrow{m^{ij}_{r}}}
\end{equation}

\vspace{-5pt}
\noindent \kk{where $\norm{\overrightarrow{m^{ij}_t}}$ and $\norm{\overrightarrow{m^{ij}_r}}$ are the Euclidean distance between $m^i$ and $m^j$ in the teach and repeat maps, respectively.}

\kk{The right-hand side of (\ref{eq-gamma}) would be zero for perfectly accurate maps without any object relocation. But in real-world applications it is nonzero, and we aim to find the best matching pair of objects, denoted as $(m^{i^*},m^{j^*})$, that minimize this aggregated error $\gamma$. That is, we search for the object indices that minimize (\ref{eq-gamma}) as follows:}
\begin{equation} 
\label{argmin}
 (i^*, j^*) = \argmin\limits_{\substack{(i,j) \in (1 \cdots N)}} \gamma(\cdots)
\end{equation}

\vspace{-5pt}
\kk{In practice, to make our relocalization algorithm more robust to estimation errors and outliers caused by object movement, we only use the top matching objects in calculating $\gamma$. Specifically, positional errors between two maps for all objects are first sorted in an ascending order. Only the top half of objects are used in (\ref{eq-gamma}) and minimizing (\ref{argmin}).}

\kk{After finding the best matching pair of objects between the two maps, the optimal scale ratio ($s^{*}$) is calculated by (\ref{eq_scale}). The optimal 3-DoF transformation matrix $T^*$ between the teach and repeat map is calculated from the best matching pair of objects ($m^{i^*},m^{j^*}$) as follows:}

\vspace{-6pt}
\begin{equation}\label{repeat_to_teach}
\text{}^{O_{t}}_{O_{r}}T^*(\delta \theta,\delta x,\delta y) = \text{}^{O_{t}}_{O^{i^*j^*}_t}T \times \text{}^{O^{i^*j^*}_r}_{O_{r}}T
\end{equation}

\vspace{-4pt}
\kk{Therefore, three of the unknowns ($\delta x, \delta y$ and $\delta\theta$) are solved for and embedded in the above matrix $T^*$. It is now straightforward to transform the robot pose in the repeat map $P_r$ to the teach map pose $\text{P}_{t}$ as follows:}

\vspace{-4pt}
\begin{equation} \label{eq_robot_trans}
 \text{P}_{t}=s^* \times (\text{}^{O_{t}}_{O_{r}}T^*\times \text{P}_{r})
\end{equation}

It is interesting to note that as long as fewer than half of the objects are moved, the object movements would be detected during the relocalization phase due to our previous assumption of only considering top matching objects. 

\mmedit{Note that while introducing object-level data association may help the algorithm to work with repetitive objects, our tests showed that due to the highly different camera poses between teach and repeat phases, the objects' side that the camera is observing can be completely different and in many cases, there may be no similarity between two sides of an object. Although training a CNN-based model using the captured images of the teach phase, similar to \cite{c28}, can help to better distinguish the objects during repeat phase, doing so is outside of the scope of this paper, as we do not save the captured images. In any case, one advantage of our algorithm is that no training is needed after the teach phase. }

\mmedit{Another solution for this problem can be to memorize the relative topological position of the objects in the scene during teach phase and use them during the repeat phase. 
However, in some cases, moving and rotating similar objects in the environment may prevent the algorithm from distinguishing objects from object-level association or relative topological positioning, which is needed in our relocalization algorithm. }

\subsection{Forward Motion}

\kk{After transforming the initial robot pose from the repeat map to the teach map, the next step is to move toward the teach path and repeat it. The forward motion involves moving toward the closest point on the teach path to the robot, and then following the list of keyframes $P^k$ captured in the teach phase toward the end. Therefore, the closest point on the teach path from the current pose of the robot is considered as the first goal point of the robot. Then the next goal points are chosen with an arbitrary distance to the current point within the range of one meter. The robot would first rotate toward the next goal point and then move from the current position toward the next one all the way to the endpoint $P^K$.}

\begin{figure}[t]
\centering
\includegraphics[scale=0.22]{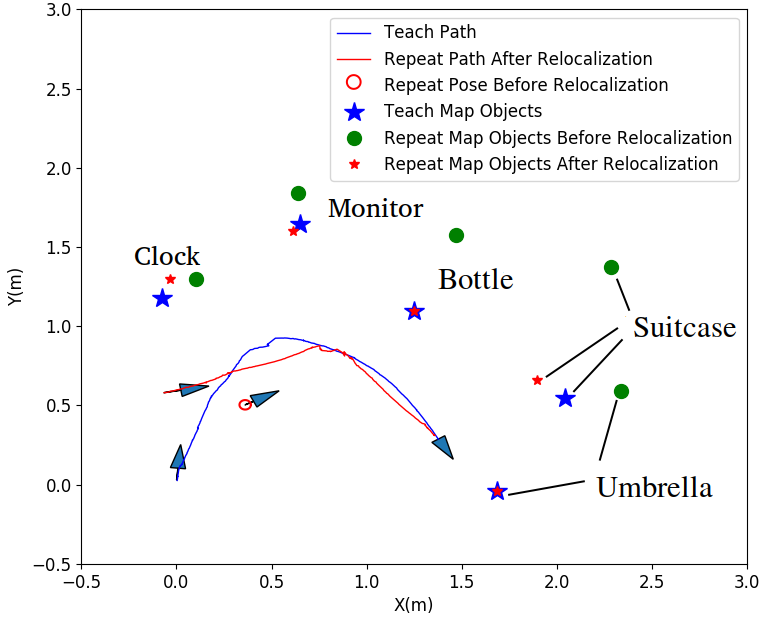} 
\caption{\kk{An example of the teach map and the repeat map before and after relocalization. Here, the bottle and the umbrella were matched, and helped to relocalize the robot pose in repeat onto the teach map. The arrows show the initial robot pose.}}
\label{teach_repeat_reloc}
\end{figure}

\subsection{Backward Motion}

\kk{The robustness of our system is manifested by its ability to repeat the teach path in a backward direction. Our VTR algorithm relocalizes the robot from the repeat map to the teach map using detected semantic landmarks, independent of the viewpoint. Then instead of choosing the next points toward the end-point on the teach path, the robot can choose points directed to the starting location on the teach map. Therefore, the backward motion is basically following the list of keyframes captured in the teach phase from the closest point on the teach path toward the starting keyframe $P^1$.}

\section{Experiments \label{sec:expt}}

\kk{To evaluate the performance and accuracy of our algorithm and system, several experiments were carried out in a lab at Simon Fraser University. The lab was equipped with a Vicon motion capture system, which was only used for evaluation purposes and not used in our VTR method. The test platform was a Turtlebot2\cite{c22} robot equipped with a ZED2 camera capturing only monocular data. The computer hardware was Intel CPU Core i9-9980HK and RTX 2080 GPU, running the ROS Kinetic on Ubuntu 16.04.}
   
\kk{The robot was placed inside the experimental area in which a number of objects were randomly placed. Then the robot was controlled manually by a user along a teach path, while ORB-SLAM, YOLOv3, and the object position detection modules were running simultaneously. The VTR system also built the semantic map and memorized the teach path in an online fashion. An example semantic map built in the teach phase can be seen in Fig.~\ref{semantic_map}.} 

\begin{figure}[t]
\centering

\subfloat{
	\label{fig:forward-first}
	\includegraphics[width=0.23\textwidth]{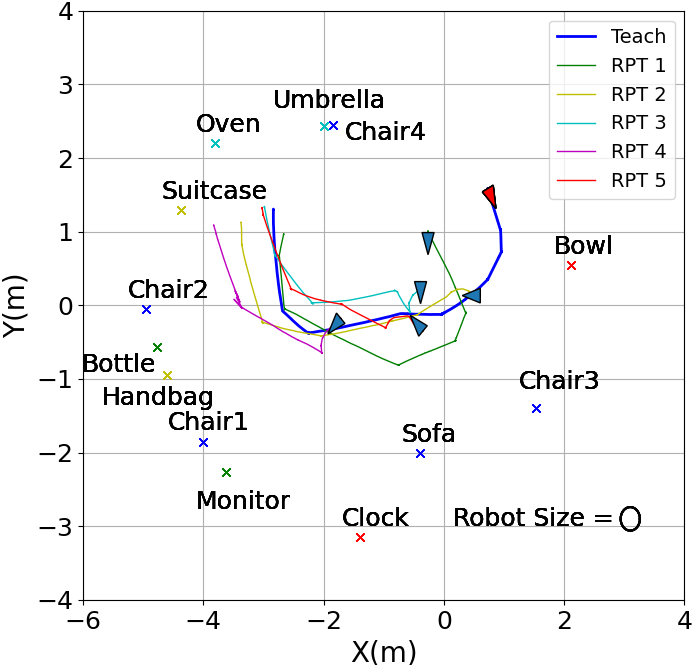} } 
\subfloat{
	\label{fig:forward-second}
	\includegraphics[width=0.23\textwidth]{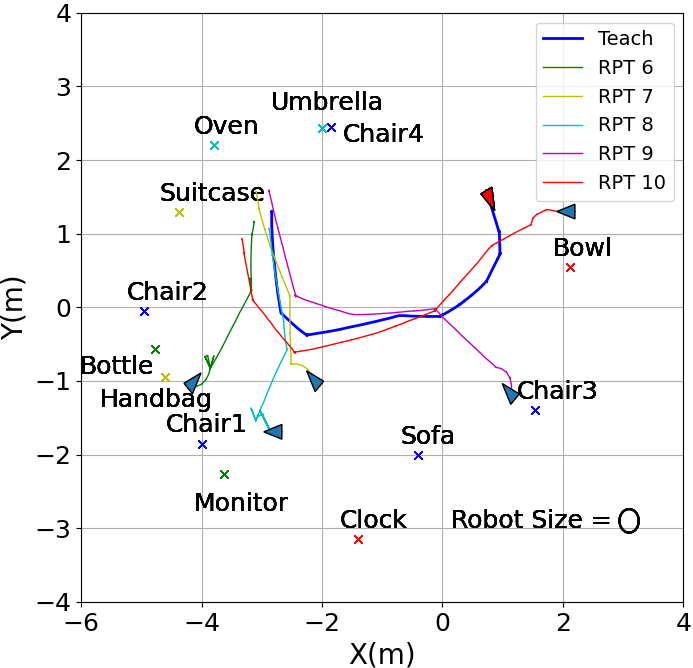}} 

\caption{\kk{The teach and ten repeat motions captured by our Vicon motion capture system for forward repeat tests. Triangle arrows indicate the initial poses of the robot, for which we chose to cover a large range of starting positions and orientations.}}
\label{teach_and_repeat}

\end{figure}

\kk{During the repeat phase, \ky{we placed the robot in various initial positions and orientations in the lab. \mmedit{Then the robot moves slightly on a pre-defined short path, either autonomously, e.g., the initial curved motion in Repeat 3, 4 and 5 in Fig.~\ref{teach_and_repeat} or manually, e.g., the initial part of the Repeat 1 and 8 in Fig.~\ref{teach_and_repeat}, to \mm{both start ORB-SLAM to track the camera pose and }briefly observe the environment for the mapping modules to create a new semantic map for the repeat phase.} The precise details of this initial motion is not too important: As long as a subset of at least three objects in the lab were seen, the relocalization module transforms the current robot pose to the teach map using the observed objects in both maps.}}

\subsection{Forward Repeat}

\kk{Having relocalized the robot with respect to the 3D semantic map built in the teach phase \mmedit{and while the ORB-SLAM is running,} the VTR system activated the motion planning module to move the robot toward the closest point on the teach path and followed its keyframes toward the end. In Fig.~\ref{teach_repeat_reloc} we show the object locations and robot paths: on the teach map; on the repeat map before relocalization; and on the repeat map after relocalization. In this specific test, the bottle and the umbrella were chosen by our algorithm to be the best matching object pair. The repeat path taken by the robot converged to the end-point with high accuracy, by traversing some of the camera keyframes of the teach path.}

\begin{figure}[t]
\centering

\subfloat{
	\label{fig:faraz}
	\includegraphics[width=0.22\textwidth]{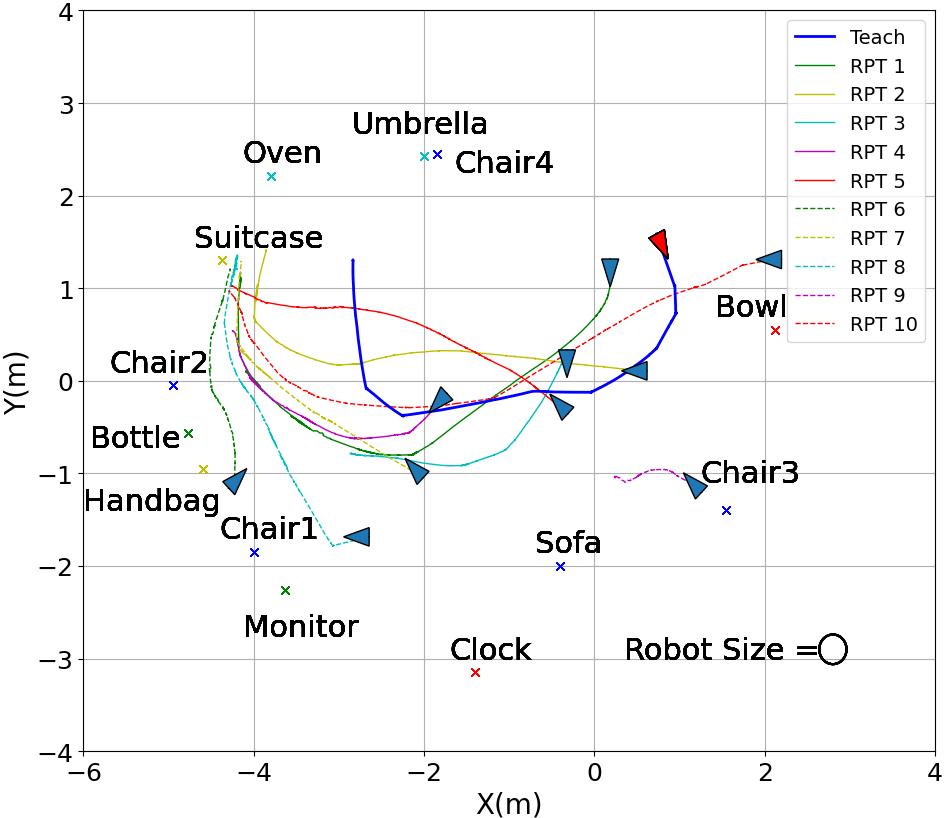} } 
\subfloat{
	\label{fig:teach_repeat}
	\includegraphics[width=0.22\textwidth]{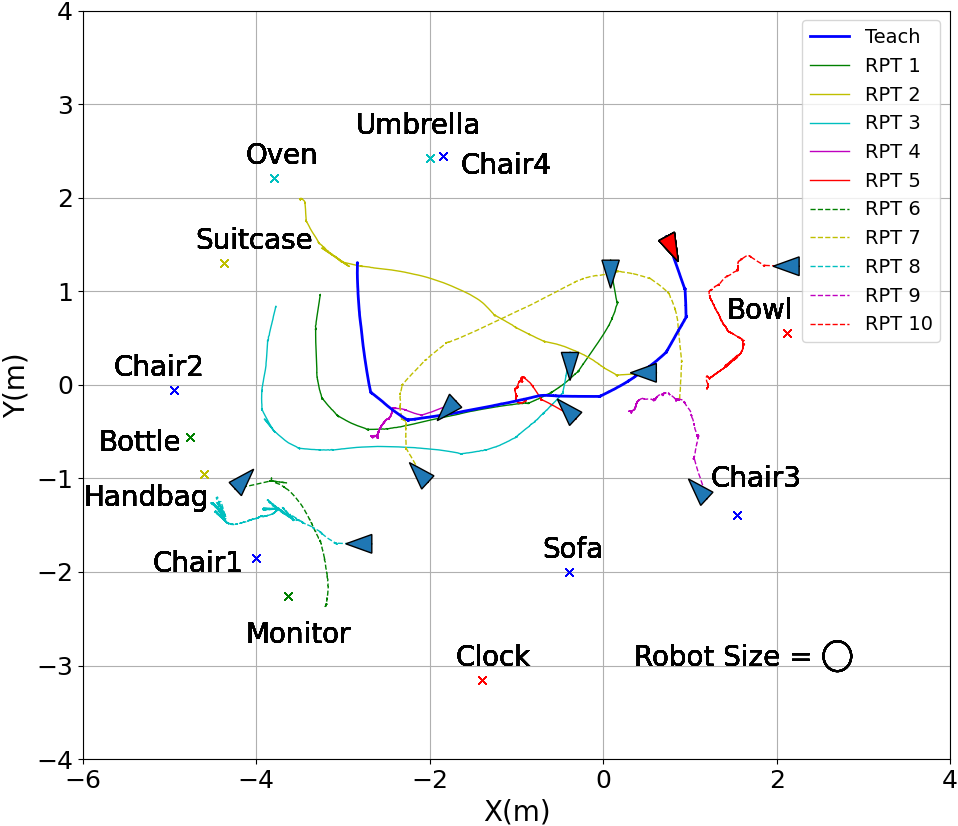}} 

\caption{\ky{The teach and repeat paths resulted from VTR algorithms published in \cite{c19}(left) and \cite{c25}(right), using similar initial robot poses as the tested poses for our own algorithm shown in Fig.~\ref{teach_and_repeat}}}
\label{evaluation_figures}

\end{figure}

\begin{table}[t]
\caption{\mmedit{Statistics of the start-point and end-point distances and angle differences between the teach path and the repeat paths. The average and standard deviation in meters and degrees are reported for both forward and backward repeats}}
\vspace{-2pt}
\label{table_1}
\begin{center}
\begin{tabular}{|c||c|}
\hline
VTR Variation of & avg. $\pm$ std. dev.\\
\hline
Start-Point Distances-Forward Motion & ${2.65} \pm {1.34 ~(\text{m})}$ \\
\hline
End-Point Distances-Forward Motion & ${0.41} \pm {0.26 ~(\text{m})}$ \\
\hline
\hline
Start-Point Distances-Backward Motion & ${1.94} \pm {1.40 ~(\text{m})}$ \\
\hline
End-Point Distances-Backward Motion & ${0.60} \pm {0.20 ~(\text{m})}$ \\
\hline
\hline
Starting Angle Differences-Forward Motion & ${\ang{155.87}} \pm {\ang{125.94}}$ \\
\hline
Ending Angle Differences-Forward Motion & ${\ang{10.50}} \pm {\ang{9.40}}$ \\
\hline
\hline
Starting Angle Differences-Backward Motion & ${\ang{149.85}} \pm {\ang{76.69}}$ \\
\hline
Ending Angle Differences-Backward Motion & ${\ang{63.09}} \pm {\ang{19.15}}$ \\
\hline
\end{tabular}
\end{center}
\end{table}

\kk{We have performed in total 10 forward repeat tests, all of which were captured by the Vicon motion-capture system and are shown in Fig.~\ref{teach_and_repeat}. The thick blue lines indicate the teach path, and all other colored lines visualize the repeat paths. The starting points for repeat motions were strategically chosen to cover a large range of locations and viewing angles inside our lab. In most cases, the viewing angles were significantly different from the one in the teach phase. The locations of the environmental objects utilized in the VTR algorithm are also shown in the figure. In all tests, the robot was first moved slightly to initialize the ORB-SLAM and semantic map for relocalization. Then the robot was able to repeat the teach path with reasonable accuracy, independent of the initial robot location and viewing angle. Also, to show our algorithm works in environments with few unique objects different tests have been performed and you can see its performance here: \url{https://www.youtube.com/watch?v=raRT7S9NSfc&list=PLuLzEWWNu1_qjPRa5OqTj6P4EjpY7veUA}}

\kk{We report key statistics of the 10 repeat tests in Table~\ref{table_1}. The distances between the start point in the teach path and the ones in the repeat paths were calculated, and the average and standard deviation in meters are given in the Table~\ref{table_1}. The end-point distances and statistics were also computed. As we can see, for forward repeats, the end points were on average almost one robot diameter (0.41 m) away from the teach path end point, despite the large range and variance of the start points in repeats. \mmedit{Also, the average and standard deviation of starting and ending robot angles for forward and backward repeats compared to the teach phase are given in the Table~\ref{table_1}. The algorithm has been able to move the robot to the ending point with an average of 10 degrees of error with respect to the teach phase.} \mm{It is important to mention, due to the scale ambiguity of generated map (between each generated semantic map and the real world) we cannot report object localization errors. Also, our focus was mainly on pose independence, that is, being able to repeat the teach path on a relatively accurate path and end up at a point closed to the teach path end-point, starting from a variety of poses. Therefore, we do not attempt to track the teach path as close as possible, which can increase the motion smoothness of the repeat path produced by our algorithm.}}

\ky{We compare the performance of our algorithm to two recent closely-related VTR algorithms with author-released code, \cite{c19} and \cite{c25}, using the same teach path in the same testing environment. The robot was placed in various locations with similar initial poses as some of the tested poses for our own algorithm shown in Fig.~\ref{teach_and_repeat}. Fig.~\ref{evaluation_figures} shows the performance of these two algorithms. In most cases, they cannot follow the teach path as well as our algorithm as shown in Fig.~\ref{teach_and_repeat}. The VTR algorithm from \cite{c19} also uses semantic objects to try to find the correct repeat path, similar to our work. However, our tests show that the robot often loses track of the teach path before it reaches the end point. Sometimes, the robot can even get stuck near the initial position and is not able to finish the test. The VTR algorithm from \cite{c25} mainly relies on recorded odometry data and tries to reduce the robot pose error using the captured image features. Not surprisingly, the robot cannot find similar image features to repeat the teach path, when there is a drastic change in the viewing angle.}
\mmledit{We do not report starting and ending position and angle errors for these methods in Table~\ref{table_1}, since they are unable to finish many tests.}

\begin{figure}[t]
\centering
\includegraphics[scale=0.22]{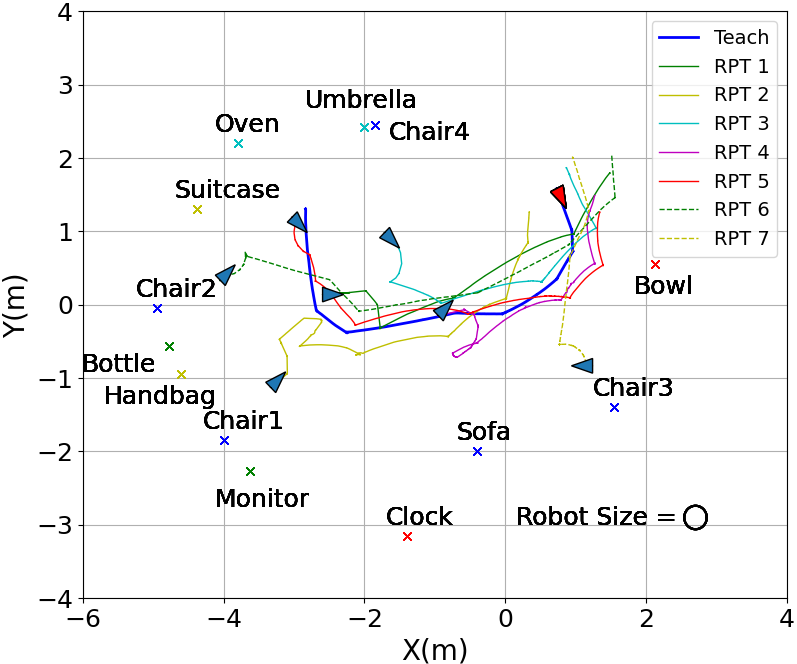} 
\caption{\kk{The teach and seven repeat motions evaluated by Vicon motion capture system for backward repeat tests. Triangle arrows indicate the initial poses of the robot, which can vary significantly such as a closed to \ang{180} viewing angle difference.}}
\label{backward_teach_and_repeat}
\end{figure}

\subsection{Backward Repeat}

\kk{We evaluate the VTR algorithm in the backward direction with the same experimental setup as in the forward repeats. The only difference is that the robot would move toward the starting point instead of the end point on the teach path. Fig.~\ref{backward_teach_and_repeat} shows the seven backward repeat motions tested in the lab. Again, we report the average and standard deviation of start-point and end-point distances and angles with respect to the teach path in Table~\ref{table_1}.}

\kk{The results indicate reasonable accuracy of our algorithm in backward repeats. In most cases, the path was followed in reasonable accuracy with respect to the size of the robot, especially considering the significant viewing angle differences, sometimes close to \ang{180}, between the teach path and the backward repeats.}

\begin{figure}[b]
\centering
\includegraphics[scale=0.22]{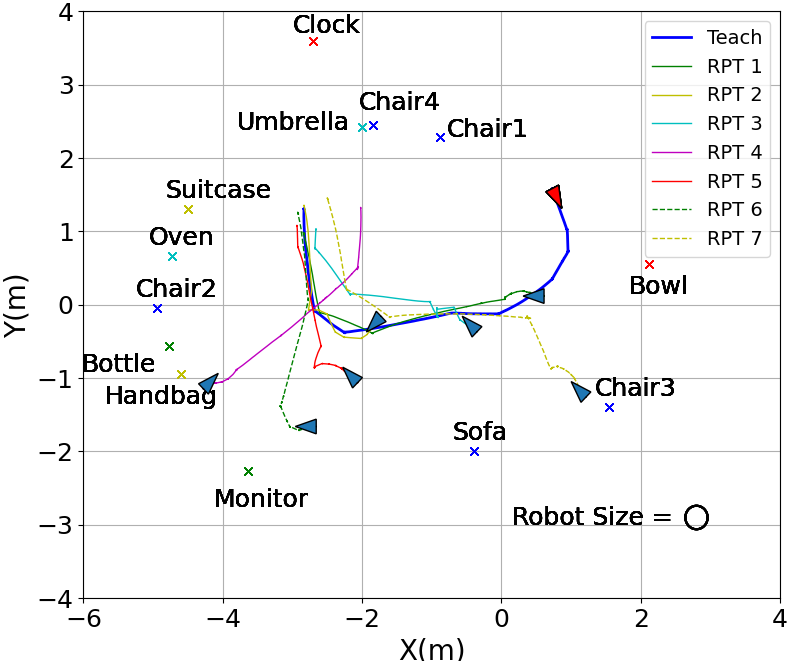} 
\caption{\kk{The teach and repeat motions captured by our Vicon motion capture system, after relocation of five objects. Old location of each object can be seen in Fig.~\ref{teach_and_repeat} and Fig.~\ref{backward_teach_and_repeat}.}}
\label{teach_and_repeat_after_movement}
\end{figure} 

\subsection{Robustness Towards Environmental Changes}
\vspace{-2pt}

\kk{We also evaluate the robustness of the proposed algorithm to occasional environmental changes. \mmedit{As a test case, we randomly moved five of the objects, i.e., a chair, handbag, clock, oven, and the suitcase that were semantic landmarks in the teach map, to the new places}. Seven more forward repeat tests were performed in the changed setting. Fig.~\ref{teach_and_repeat_after_movement} shows the teach and repeat paths as well as the new positions of all landmarks. The repeat motions were as accurate as before qualitatively, as can be seen in Fig.~\ref{teach_and_repeat_after_movement}, and quantitatively too, achieving comparable performance as reported in Table~\ref{table_1}. 
\ky{Similarly, other objects could be moved or removed from the environment without degrading the repeat path. Our VTR is robust to multiple environmental changes as long as the number of relocated objects is less than half of all the semantic objects, \mmedit{ as our algorithm only uses the top half of objects with the lowest positional error to relocalize as explained in section~\ref{sec:relocalization}.}} This algorithm feature holds for backward repeats as well.}

\subsection{Comparison with Image-Based Algorithms}
\label{sec:comparison}

\begin{figure}[t]
\centering
\includegraphics[scale=0.2]{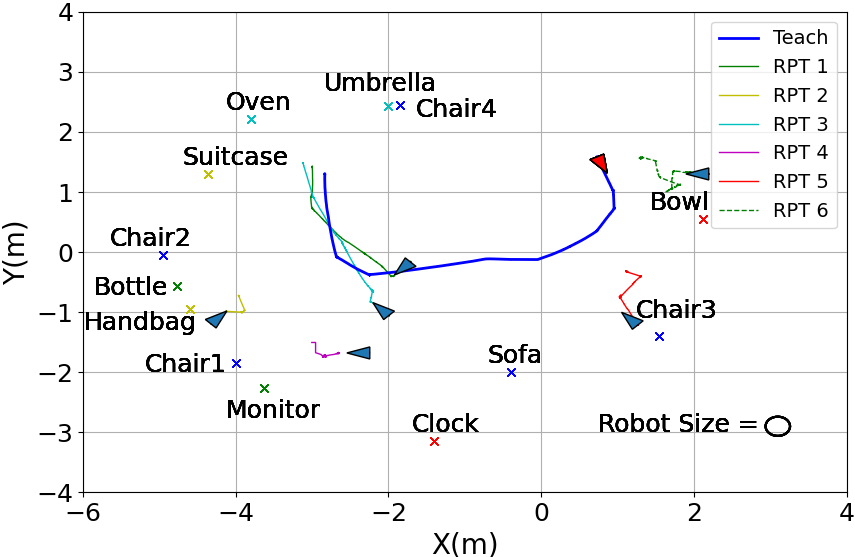} 
\caption{\mmedit{The teach and repeat motions captured by our Vicon motion capture system, while purely using ORB-SLAM for camera pose relocalization.}}
\label{teach_repeat_ORB}
\end{figure}

\mmedit{In order to demonstrate the advantages of our proposed algorithm, we compared our method with two image-based localization algorithms. In the first case, we tested the relocalization capability of only using ORB-SLAM~\cite{c6} as a visual SLAM method. Previously, reference \cite{c5} had shown ORB-SLAM is not capable of relocalizing the camera inside a created map beyond 30 degrees due to the lack of feature matches. To validate this again, we moved the robot through the teach path while using ORB-SLAM to create a map. Then, we moved the robot to each one of the repeat paths initial points and slightly moved and rotated the robot to re-initialize the ORB-SLAM. In most of the cases, where the camera pose was highly different with respect to the teach path, the ORB-SLAM was not able to relocalize the camera. However, in some of the cases with similar camera poses between teach and repeat phases, the ORB-SLAM could relocalize the camera position and the robot moved toward the end-point. Fig.~\ref{teach_repeat_ORB}}

\begin{figure}[t]
\centering
\includegraphics[scale=0.20]{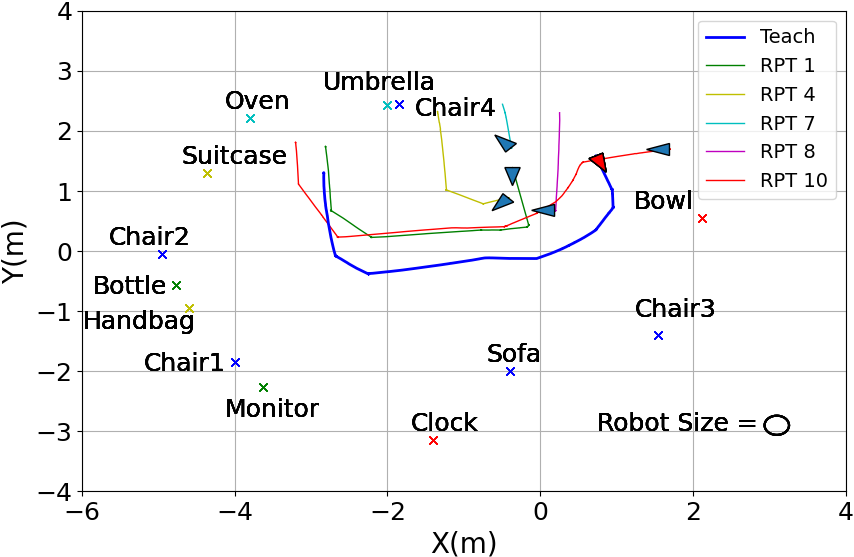}
\caption{\mmedit{The teach and repeat motions captured by our Vicon motion capture system, after relocalizing the robot between teach and repeat initial points by combining Superglue and RANSAC algorithms.}}
\label{teach_repeat_superglue}
\end{figure}

\mmedit{Also, we compared our method with a relocalization algorithm combining Superglue~\cite{c26} and RANSAC~\cite{c27}. In order to help the algorithm, we rotated the camera toward the area that the camera is looking at the teach phase initial pose. Therefore, there would be more similarity between teach and repeat phases initial poses. After calculating the transformation by superglue+RANSAC, we moved the robot to the estimated point and performed the repeat motion. The robot used the odometry data to move toward the end-point. We noticed this relocalization algorithm works accurately when the teach and repeat phases initial points are not far from each other. Fig.~\ref{teach_repeat_superglue} shows some of the repeat motions after relocalizing by superglue+RANSAC. As one can see, the repeat initial points that are close to the teach initial point, i.e, Repeat1 and Repeat10, could repeat the teach path accurately. But the other points performed the repeat motion poorly. The equivalent point for each repeat initial point and path can be found in Fig.~\ref{teach_and_repeat}.} \mmledit{We omit these methods from  Table~\ref{table_1} since they do not complete many of the tests.}

\subsection{Memory Requirement and Computation Time Comparisons}

\mmedit{We measured the memory requirement of our algorithm and compared it with \cite{c19},\cite{c25} and pure ORB-SLAM. \mmledit{We measured the peak memory usage and computation time of the whole algorithm. Our method used never used more than} 1.6 GB of memory. This number was 1.8 GB for \cite{c19} and 40 GB for \cite{c25} as they store all the images of the teach path. Also, the Superglue+RANSAC method required 1.7 GB. Pure ORB-SLAM required 0.6 GB for the cost of much lower performance as Fig.\ref{teach_repeat_ORB} shows. Compared to ORB-SLAM, our method requires more memory for the object detection module, which is required for the robustness and pose-independence of the VTR algorithm. We also measured the maximum computation time, over all execution time steps, of each method. References \cite{c19},\cite{c25}, Superglue+RANSAC, pure ORB-SLAM and our method required 0.001, 0.1, 0.1, 0.003 and 0.03 seconds respectively. Our python implementation is not optimized, but still fast enough for running our algorithm in real-time.}

\section{Conclusion, Discussion and Future Work}

\kk{We have proposed a novel VTR algorithm using 3D semantic maps. The algorithm is robust to large robot starting pose changes in the repeat phase. Our key insight is to build 3D semantic maps, containing semantic labels and positions of the objects in the environment as well as camera poses, during the teach phase. In the repeat phase, the algorithm relocalizes the robot in the new map based on the found objects. We tested our method in both forward and backward modes starting from various locations inside the lab. Our algorithm demonstrated robustness toward significant starting pose variations, as well as tolerance to environmental changes. \ky{Our algorithm compares favorably with respect to two other most-relevant recent VTR methods \mmedit{and two image-based relocalization algorithms} in terms of both accuracy and robustness.}}

Although we were able to show a simple semantic map can help to relocalize the robot and perform an accurate, robust, and pose-independent VTR, there are still multiple areas for future research. For example, we showed the performance of our algorithm in an indoor area. While it was not the main focus of this paper, it would be interesting to test our algorithm for repeating for much longer and possibly outdoor paths. Also, the accuracy of the semantic maps could be improved by a more sophisticated 3D position estimation algorithm. In certain environments with very few objects, it may be useful to utilize repetitive objects in the environment in the future. We also wish to improve the end-point accuracy for applications that require return trips and loops, and consider environments with other mobile agents.

\end{document}